%% file: main.tex
\documentclass{article}

\usepackage[preprint,nonatbib]{neurips_2020}

\usepackage[utf8]{inputenc} % allow utf-8 input
\usepackage[T1]{fontenc}    % use 8-bit T1 fonts
\usepackage{hyperref}       % hyperlinks
\usepackage{url}            % simple URL typesetting
\usepackage{booktabs}       % professional-quality tables
\usepackage{amsfonts}       % blackboard math symbols
\usepackage{nicefrac}       % compact symbols for 1/2, etc.
\usepackage{microtype}      % microtypography
\usepackage{graphicx}       % for includegraphics
\usepackage{bm}
\usepackage{amsmath}

\renewcommand{\vec}[1]{\mathbf{#1}}
\newcommand{\mat}[1]{\mathrm{\mathbf{#1}}}
\newcommand{\graph}[1]{\mathcal{#1}}
\newcommand{\etal}{\textit{et al.}}

\title{GraphMDN: Leveraging graph structure and deep learning to solve inverse problems}

\author{%
  Tuomas Oikarinen\\
  Massachusetts Institute of Technology\\
  Cambridge, MA 02139, USA\\
  \texttt{tuomas@mit.edu} \\
  \And
  Daniel C. Hannah\\
  Vectra AI\\
  San Jose, CA 95128, USA\\
  \texttt{dhannah@vectra.ai}
  \And
  Sohrob Kazerounian\\
  Vectra AI\\
  San Jose, CA 95128, USA\\
  \texttt{sohrob@vectra.ai}
}

\begin{document}

\maketitle

\begin{abstract}
    The recent introduction of Graph Neural Networks (GNNs) and their growing popularity in the past few years has enabled the application of deep learning algorithms to non-Euclidean, graph-structured data. GNNs have achieved state-of-the-art results across an impressive array of graph-based machine learning problems. Nevertheless, despite their rapid pace of development, much of the work on GNNs has focused on graph classification and embedding techniques, largely ignoring regression tasks over graph data. In this paper, we develop a Graph Mixture Density Network (GraphMDN), which combines graph neural networks with mixture density network (MDN) outputs. By combining these techniques, GraphMDNs have the advantage of naturally being able to incorporate graph structured information into a neural architecture, as well as the ability to model multi-modal regression targets. As such, GraphMDNs are designed to excel on regression tasks wherein the data are graph structured, and target statistics are better represented by mixtures of densities rather than singular values (so-called ``inverse problems"). To demonstrate this, we extend an existing GNN architecture known as Semantic GCN (SemGCN) to a GraphMDN structure, and show results from the Human3.6M pose estimation task. The extended model consistently outperforms both GCN and MDN architectures on their own, with a comparable number of parameters. 
\end{abstract}

\input{1_introduction}

\input{2_background}

\input{3_methods}
\input{4_results}

\section{Conclusion}

   We have introduced GraphMDN, a novel neural network architecture that combines notions from semantic graph convolutional networks and mixture density networks. Our primary aim in this paper is to introduce GraphMDN and demonstrate its general utility on a common and important dataset, where we attain noticeable performance improvements without any significant pose-estimation-specific adaptations. Indeed, pose estimation is but one potential application of the GraphMDN. Any graph-structured inverse problem (such as predicting distributions along edges or nodes) should benefit from our novel network architecture allowing a mixture density network to operate over graph-structured input data.

{\small
\bibliographystyle{ieee_fullname}
\bibliography{egbib}
}

\end{document}

%% file: 1_introduction.tex
\section{Introduction}\label{introduction}

Despite the improvements brought about by deep learning in problem domains spanning speech and language to computer vision and control, neural network architectures were largely incapable of operating on non-Euclidean, graph-structured data. This shortcoming is particularly relevant in light of the importance and ever-increasing availability of data that is intrinsically graph-structured; examples include social networks, protein structures, molecules, knowledge graphs, and computer networks. It is only with the recent introduction of Graph Neural Networks (GNNs) \cite{bronstein2017geometric,scarselli2008graph} that deep learning techniques have been extended to directly operate over graph-structured data. By drawing inspiration from the mathematical tools that allowed traditional deep learning algorithms to excel at learning from data such as images and speech, GNNs have begun to make use of operators like graph convolutions, enabling them to operate over non-Euclidean graph structured inputs in a manner similar to standard convolutions over images. Indeed, as pointed out by Bronstein \etal \cite{bronstein2017geometric}, traditional Convolutional Neural Networks can be thought of as operating over graph structured data, where images are simply graphs sampled over a two-dimensional grid.

By combining principles from deep learning with the ability to operate over a much broader range of data, GNNs have been able to achieve impressive results on a wide range of tasks historically considered out-of-reach for traditional deep learning models (e.g., relation prediction in knowledge graphs \cite{nathani2019learning,hamilton2018embedding}, drug design and interaction prediction \cite{zitnik2018modeling}, visual and language-based recommender systems \cite{ying2018graph}, binary-code analysis \cite{xu2017neural,baldoni2018unsupervised}, etc.) Much of the focus of GNN research has been on architectures for node classification, link prediction, and clustering/embedding graph structures. Regression tasks have received less attention. In the context of graphs, the goal of a regression task is to predict some continuous-valued output at the graph, node, or edge level (or some combination thereof). While GNNs are an obvious choice for any regression task where the inputs are graph structured, they face the same difficulties as standard deep-learning architectures when predicting continuous valued outputs. If the target statistics of an output given some input are not unimodal (i.e. the mapping to be learned by the model is one-to-many, a so-called ``inverse" problem), a standard output layer with a mean-squared error loss function will tend to learn the conditional average of the outputs observed for a given input \cite{bishop1994mixture}. As a simple example, consider a neural network tasked with predicting the amount of traffic uploaded to a server from a machine on a network. 

If the amount of traffic to be predicted given some input X is bimodal (e.g,. half the time uploading around 10 GB, and the other half uploading 100 GB of data), a standard neural network learning the regression target will tend towards simply predicting the mean of these values. While the conditional average may be useful in some cases, a model predicting that the machine uploads 55 GB does not fully characterize the upload behavior of this example machine.

Motivated by the use of GNNs in cybersecurity, where it is often critical that models are able to predict multi-valued quantities at the edge (e.g. bytes transferred between two hosts) and node (e.g. number of login attempts on a host in any given day) level, we introduce Graph Mixture Density Networks (GraphMDNs). GraphMDNs combine the ability of GNNs to operate over graph-structured data with the representational power of mixture density networks in modeling the statistical distributions of target data. In the present work, we demonstrate the general utility of this architecture by applying it to a common dataset for 3D human pose estimation from monocular images or 2D joint positions, and show that GraphMDNs exceed the capabilities of either a GNN or an MDN on their own with minimal changes to the architecture. Human pose estimation serves as an ideal platform with which to benchmark GraphMDN due to availability of large high quality datasets, and the fact that graph convolutional networks (GCN) and mixture density networks have separately been applied to this problem in prior works. The next section provides an overview of graph neural networks, mixture density networks and previous work in human pose estimation. Section \ref{architecture} describes the architecture of GraphMDNs. Experiments on the Human3.6M dataset are discussed in section \ref{experiments}, with results and comparison to other models presented in section \ref{results}.

%% file: 2_background.tex
\section{Background}\label{background}

\subsection{Graph neural networks}\label{gnn_background}
Graph Neural Networks (GNNs) (introduced in \cite{gori2005new} and \cite{scarselli2008graph}) have extended the capabilities of neural network and deep learning models to the graph domain, and have been used for node, edge, and graph level prediction and embedding (see \cite{wu2019comprehensive,zhou2018graph} for recent, comprehensive reviews of the state of GNNs). Prior to GNNs, applying standard neural network architectures to graph-structured data required an unwieldy degree of hand-engineered features in order to make the data suitable for training, obviating many of the benefits that motivate the use of deep learning in the first place. By extending to the graph domain operations such as convolution, which has proved indispensable in deep learning for computer vision, GNNs enable deep learning models to represent, learn and make predictions from topological structures that they would have otherwise been incapable of processing.

Roughly speaking, the extension of neural networks to graphs can be thought of as falling into either spectral or non-spectral methods. In the former, filters (e.g., spectral graph convolutions) are defined to operate in the Fourier space of the graph \cite{defferrard2016convolutional,bruna2013spectral}, whereas in the latter, the convolution is defined to operate on the graph directly, operating over a spatial neighborhood and with a set of shared parameters \cite{duvenaud2015convolutional,hamilton2017inductive}. While each approach has advantages and dis-advantages (e.g., spectral methods often require computation of the graph Laplacian, which can be computationally expensive) --- the framework we introduce here combining mixture density networks and graph neural networks, can operate on either. The particular GraphMDN we demonstrate here builds on the Semantic Graph Convolutional Network (SemGCN; \cite{zhao2019semantic}) model for pose estimation and falls into the non-spectral approach. 

\subsection{Mixture density networks}\label{mdn_background}

Mixture density networks (MDNs) were introduced by Bishop \cite{bishop1994mixture} in order to overcome limitations in the representational capacity of traditional neural networks. In particular, Bishop demonstrated that in either regression or classification, neural networks learn to approximate the conditional average of the target data given an input. While this will often suffice in classification tasks, and can even be considered optimal (since the goal of the network is to explicitly model the posterior probability for class membership of an input), it can lead to unsatisfactory results in the case of regression tasks. This is due to the fact that the conditional average of a particular regression target can range from insufficient to misleading (as in the computer network example given in Section \ref{introduction}).

To overcome this problem, Bishop modified the traditional neural network architecture so that, instead of single scalar values, the network would output the parameters of a mixture of distributions. This can represent the more complex targets observed for inverse problems and other datasets where a single input results in multiple distinct outcomes. If the mixture is comprised solely of Gaussian distributions (this need not be the case in general), the conversion of a standard neural network to a MDN can be achieved by replacing the neuron(s) in the output layer with a set of $3N$ neurons, where $N$ is the number of Gaussian distributions in the output mixture. Of the $3N$ neurons, $2N$ model the mean and variance, respectively, of each of the $N$ distributions, along with $N$ neurons that output the mixing coefficients of those distributions. Because MDNs leverage the function-approximation capabilities of neural networks to learn parameterizations of mixture models, they go beyond standard mixture models insofar as their input domains are not restricted to continuous-valued inputs, or even static data. MDNs can trivially incorporate categorical features and timeseries data as input features, while still parameterizing a mixture of distributions on the output side. Futhermore, standard mixture models define a single mixture of distributions over the whole input domain. On the other hand, MDNs learn a family of mixtures over the input domain, because for any given input, the network outputs a new parameterization and hence, a new mixture model.

\subsection{Human pose estimation}\label{human_pose_background}
2D to 3D Human Pose estimation has been well studied in the recent years. The goal of the field is to estimate a human 3D pose (location of a set of joints) based on an input image. While many approaches do this end-to-end, Martinez \etal \cite{martinez2017simple} proposed splitting this task into detecting 2D joint locations from an image, and then estimating 3D pose based on the 2D joint detections. This approach has proved effective and the focus of our work is on the second step of this pipeline, predicting the 3D joint locations based on 2D joint detections of some model. Martinez \etal 
\cite{martinez2017simple} use a fully connected network for this second step, which performs well, but does not take advantage of the inherent graph structure in the human skeleton. Recent approaches \cite{Zhao_2019_CVPR, CI_2019_ICCV} have successfully applied graph convolution networks (GCN) to improve over fully connected approaches. As the SemGCN \cite{Zhao_2019_CVPR} architecture performs well on this pose estimation task and is supported by a high-quality code release, we build on it in this paper. SemGCN is described in more detail in the next section.

Recently, algorithms have been presented that outperform SemGCN and related methods by effectively utilizing additional information available in human pose estimation datasets. For example Iskakov \etal \cite{Iskakov_triangulation} propose a geometry-inspired method that can reach very high accuracy by combining information from multiple camera views of the same action. In addition, several works \cite{rayat2018exploiting, pavllo20193d, cheng20203d} have proposed using a time-series of input frames to reduce the inherent ambiguity of the problem by analyzing how the pose evolves over time. Assuming the information is available, both of these methods could be integrated with GraphMDN for additional improvement (a point we discuss in \ref{results}), but our focus in this report is on methods that analyze a single camera and a single frame, a common scenario for real-world use cases.

\subsection{SemGCN}
Graph convolutional networks naturally operate on graph structured data. Suppose $\graph{G} = \{\mat{V}, \mat{E}\}$ is a graph given by the set of $K$ nodes $\mat{V}$ and edges $\mat{E}$. Standard graph convolutional networks share a learned kernel matrix across all edges in the graph; as a result, the ostensibly-variable relationships between nodes in the graph are not exploited as well as they could be. Semantic Graph Convolutional Networks address this issue by adding to this convolution operation a learnable weighting matrix that can account for differences in the relationships represented by the edges.

Like the original SemGCN work, we also make use of a non-local layer \cite{buades2005non, wang2018non} to capture global semantic relations between nodes. See the original SemGCN paper \cite{Zhao_2019_CVPR} for details of the non-local layer as well as the GCN structure used.

A schematic of our network architecture is displayed in Figure 1. Initially, a semantic graph convolution layer interleaved with a non-local layer is used to convert a set of 2D joint data (and accompanying adjacency matrix) to a latent representation. The latent embeddings are fed to a block structure which is repeated four times and consists of two semantic convolutional layers, each of which is followed by a batch normalization layer and a ReLU activation layer.

%% file: 3_methods.tex
\section{Architecture}\label{architecture}

\begin{figure*}[h]
  \begin{center}
  %\framebox[4.0in]{$\;$}
  \includegraphics[width=\linewidth]{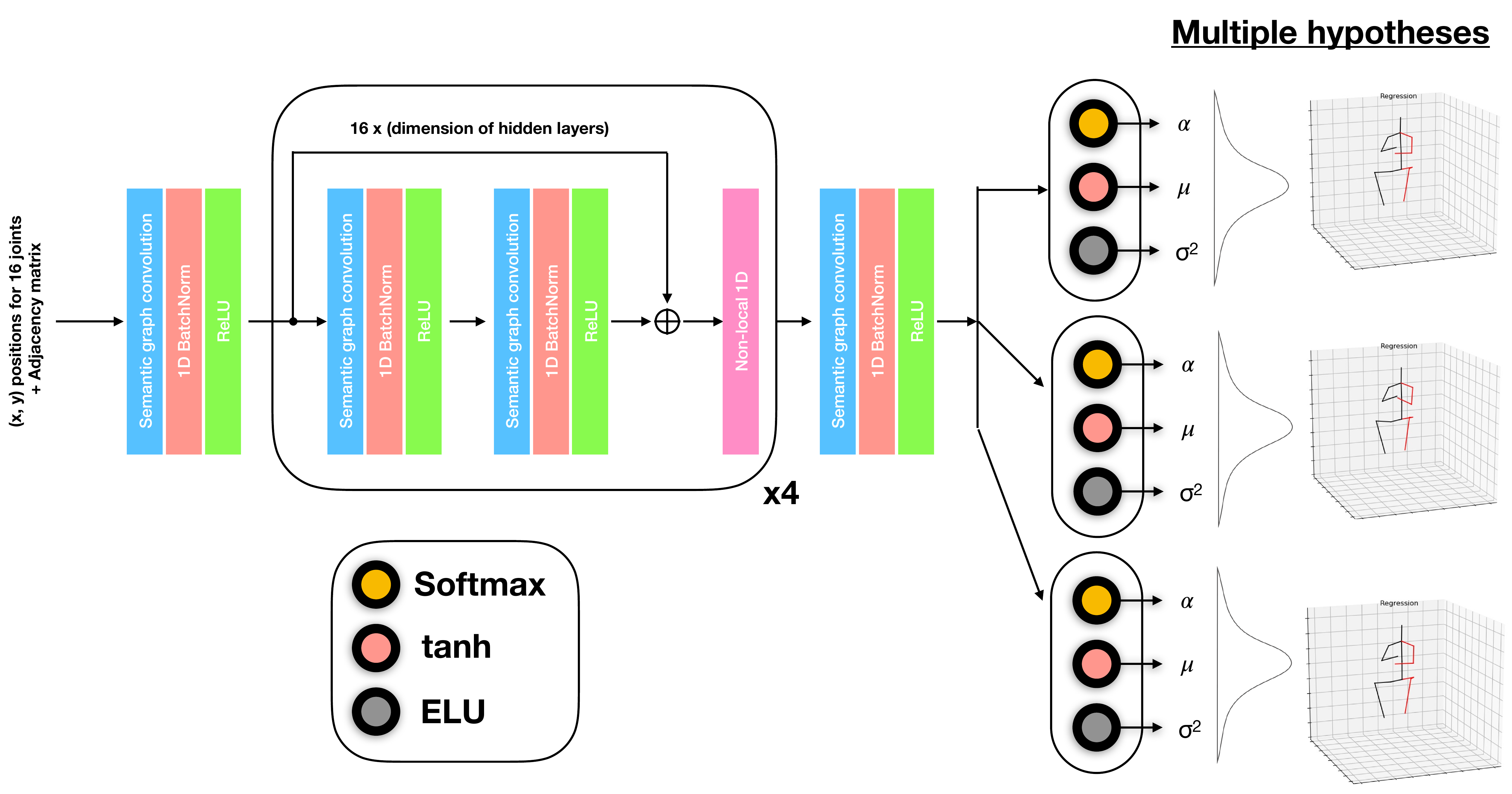}
  \end{center}
  \label{fig:arch_figure}
  \caption{Example of the GraphMDN architecture. The fundamental GraphMDN building block is a residual block (two semantic graph convolutional layers each followed by batch norm and ReLU activation) followed by a non-local layer. The fundamental building block is repeated four times and fed into a mixture density network, which generates the multiple pose hypotheses.}
\end{figure*}

Fundamentally, our neural network architecture combines Semantic Graph Convolutional Networks (SemGCN) \cite{Zhao_2019_CVPR} with Mixture Density Networks \cite{bishop1994mixture} and a method of going from node/joint-level outputs to graph/pose-level distributions.          

Motivated by the fact that multiple 3D poses may map onto the same 2D pose, our output architecture expands upon the SemGCN by employing a mixture density network (MDN). Rather than mapping a pose (represented as a graph input) $\vec{x} \in \mathbb{R}^{2K}$ to a single output $\vec{y} \in \mathbb{R}^{3K}$, we instead learn a probability density function in $\mathcal{P}:\mathbb{R}^{2K} \times \mathbb{R}^{3K} 
\rightarrow \mathbb{R}$. This is learned in the form of a mixture density network, where the SemGCN backbone with learnable parameters $w$ learns to output parameters $\Theta$ of a Gaussian mixture as a function of 2D inputs $\vec{x}$. 

\subsection{Node level probability distributions}

Since the outputs of SemGCN are node level features, the natural approach would be to learn node level distributions $\{p^1(\vec{y}^1|x), ..., p^K(\vec{y}^K|x)\}$ where $p^i$ is the probability of a 3D joint position given a 2D input and 
\begin{equation}
\mathcal{P}(\vec{y}|x) = \prod_{i=1}^{K} p^i(y^i|x). 
\end{equation}
Each $p^i$ could then be constructed via node-level Gaussian mixtures parameterized by $\Theta^i(x, w) = \left\{\boldsymbol{\mu}^i, \boldsymbol{\sigma}^i, \boldsymbol{\pi}^i\right\}$ for node i. $\boldsymbol{\mu}^i, \boldsymbol{\sigma}^i, \boldsymbol{\pi}^i$ are the means, variances, and mixing coefficients of the mixture model defined as follows:
\begin{equation*}
\boldsymbol{\mu}^i = \left\{\mu_1^i, ..., \mu_M^i | \mu_j^i \in \mathbb{R}^{3}\right\},
\boldsymbol{\sigma^i} = \left\{\sigma_1^i, ..., \sigma_M^i | \sigma_j^i \in \mathbb{R}\right\},
\end{equation*}
\begin{equation}
\boldsymbol{\pi}^i = \left\{\pi_1^i, ..., \pi_M^i | 0 \leq \pi_j^i \leq 1, \sum_j \pi_j^i = 1\right\}
\end{equation}

Where superscript $^i$ denotes the $i$th node and subscript $_j$ represents the $j$th kernel, $M$ denotes the number of Gaussian kernels and $w$ are the learnable parameters of the model. 
In SemGCN\cite{Zhao_2019_CVPR} the output of each node is a 3-dimensional prediction of the joint location. In our work this is replaced by $5M$-dimensional output of the mixture parameters $\Theta^i$. Each of the mixture components has a different type of activation function: a $\mathrm{tanh}$ layer for the means ($\boldsymbol{\mu^i}$) since targets in this task are in $[-1,1]$, a softmax for the mixture coefficient ($\boldsymbol{\pi^i}$) to ensure that it sums to 1, and an ELU +1 function for the variances ($\boldsymbol{\sigma^i}$) to ensure they remain positive. Using these parameters we can calculate the node level probability density for the ith joint being in location $\vec{y}^i$: 
\begin{equation}
p^i(\vec{y}^i|x) = \sum_{j=1}^{M} \pi_j^i f(\vec{y}^i \mid \mu_j^i, \sigma_j^i)
\end{equation}

Here, $f(\vec{y}|\mu^i_j, \sigma^i_j)$ is the probability density function of a multivariate Gaussian distribution having mean $\mu^i_j$ and a diagonal covariance matrix with diagonal elements $\sigma^i_j$, shown explicitly in Equation \ref{eq:gaussian}. The diagonal covariance matrix forces the $x$, $y$, and $z$ dimensions to be independent, but with a shared variance. In our experiments, allowing the model to freely learn a covariance matrix reduced performance. We hypothesize that the reduced performance stems from a loss of regularization provided by the restricted covariance matrix.

\begin{equation} \label{eq:gaussian}
  f(\vec{y}^i \mid \mu_j^i, \sigma_j^i) = \frac{1}{(2\pi)^{3/2}(\sigma_j^i)^3} \exp{-\frac{\|\vec{y}^i-\mu_j^i\|^2}{2(\sigma_j^i)^2}}.
\end{equation}

Training of the neural network is achieved by maximizing the log posterior probability of the observed samples, a standard practice for MDNs. The loss function is shown below

\begin{equation} \label{eq:loss_fun}
  L = -\ln{\mathcal{P}(\vec{y}|x, w)} =  -\sum_{i=1}^{K}\ln{\sum_{j=1}^{M} \pi_j^i f(\vec{y}^i \mid \mu_j^i, \sigma_j^i)}
\end{equation}

This is then minimized with respect to trainable parameters $w$. We note that it is essential to use the log-sum-exp trick when evaluating the inner sum to avoid numerical issues.

\subsection{Pose level distributions}

While the node level probability distributions are a natural representation for the task, they are restricted to learning independent distributions for each node. This is problematic when predicting realistic full body poses, because some combinations of joint positions are much more likely than others. This can be overcome by using graph/pose-level mixture coefficients $\boldsymbol{\pi}$. This transforms the problem from modeling K 3-dimensional distributions to modeling one 3K-dimensional distribution. While we still model the different dimensions as having 0-covariance, now each of the M kernels represents a plausible 3d pose, and their means can be used for creating M possible predictions.

Pose-level outputs can be achieved in several ways. The most straightforward calculation entails simply averaging the logits of each node corresponding to mixture coefficients $\boldsymbol{\pi}^i$ before applying softmax, which we found to be effective and is the approach taken when generating the results presented here. We also explored utilizing an additional dense layer to produce pose-level outputs, but found that it made training less stable and prone to overfitting. Additionally we found it beneficial to restrict the pose level distributions to have uniform variance by similarly averaging the logits of each node corresponding to $\boldsymbol{\sigma}$. The outputs for $\boldsymbol{\mu}$ are created by concatenating the outputs of different nodes.

The conceptual shift to pose level distributions also changes the way probabilities are calculated. The new model outputs are:

\begin{center}
    \begin{equation*}
    \boldsymbol{\mu} = \left\{\mu_1, ..., \mu_M | \mu_j \in \mathbb{R}^{3K}\right\},
    \end{equation*}
\end{center}

\begin{center}
    \begin{equation*}
    \boldsymbol{\sigma} = \left\{\sigma_1, ..., \sigma_M | \sigma_j \in \mathbb{R}\right\},
    \end{equation*}
\end{center}

\begin{equation}
\boldsymbol{\pi} = \left\{\pi_1, ..., \pi_M | 0 \leq \pi_j \leq 1, \sum_j \pi_j = 1\right\}
\end{equation}

And we calculate the probabilities as follows:

\begin{equation}
\mathcal{P}(\vec{y}|x) = \sum_{j=1}^{K} \pi_j(y|x)g(\vec{y} \mid \mu_j, \sigma_j). 
\end{equation}

Where:
\begin{equation} \label{eq:gaussian_2}
  g(\vec{y} \mid \mu_j, \sigma_j) = \frac{1}{(2\pi)^{3K/2}(\sigma_j)^K} \exp{-\frac{\|\vec{y}-\mu_j\|^2}{2(\sigma_j)^2}}.
\end{equation}

This is again optimized by minimizing:
\begin{equation} \label{eq:pose_loss_fn}
  L = -\ln{\mathcal{P}(\vec{y}|x, w)} =  -\ln{\sum_{j=1}^{K} \pi_j(\vec{y}|x)g(\vec{y} \mid \mu_j, \sigma_j)}
\end{equation}

Since we found pose level distributions to improve performance according to metrics used in previous works, for simplicity we only report results using pose level distributions in the rest of this paper. 

%% file: 4_results.tex
\section{Experiments}\label{experiments}

In this section, we first describe the details of our data set, then we describe the implementation details of our evaluation and training scheme. Finally we discuss how we select a hypothesis from the set of possible poses created by the MDN when evaluating our method.

\subsection{Dataset}\label{experiments_dataset}

   We evaluate our performance on the Human3.6M dataset \cite{ionescu2013human3}, which contains 3.6 million images attained by a motion capture system with four synchronized cameras in an indoor environment. The images themselves consist of seven professional actors performing daily-life activities such as walking, eating, sitting, smoking, and engaging in a discussion. Both 2D and 3D ground-truth joint locations are available, with the ultimate goal of learning to predict the 3D positions for each joint, given either the raw image, or the 2D ground truth locations of the joints in that image. 
   
   When predicting the 3D joint positions from a single 2D input (i.e., using inputs that are only a single camera view and a single moment in time), the task of determining the 3D joint locations is underdetermined. For example, movement in the depth-axis of the camera and occlusion of joints in a particular camera view create scenarios where many 3D joint positions and poses can be projected onto the same 2D input view. As such, state-of-the-art techniques now often sidestep the difficulties resulting from the ``inverse problem'' nature of the task by making use of the synchronized multi-view images, or the full temporal sequence of images for each action/subject. 
   
   In order to demonstrate how a GNN and MDN combination extends representational capacity beyond either the GNN or MDN model alone, we show results in the case of single-view/single-frame prediction given a single 2D input. We first compare results to the SemGCN model utilizing ground-truth 2D inputs for training and testing (hereafter denoted GT), and second, compare to a multi-modal MDN model using fine-tuned stacked hourglass 2D detections given by \cite{martinez2017simple} as an input (hereafter denoted SH). We then qualitatively show the utility in a model that can entertain and represent multiple hypotheses simultaneously, and discuss further benefits that can be derived from such a representation by incorporating multi-view or temporal information in order to more optimally select between them. 

\subsection{Training and evaluation protocols}\label{experiments_training}

   Following previous work \cite{Zhao_2019_CVPR, pavllo20193d, jahangiri2017generating, Li_2019_CVPR}, we train on subjects 1, 5, 6, 7, and 8 and test on subjects 9 and 11. We use all the images from all four camera views in training and testing, albeit not in a multi-view setup; the different camera views are treated as independent examples. 

   To evaluate our network, we consider two commonly used protocols for evaluating the performance of 3D pose estimation, henceforth referred to as Protocol \#1 and Protocol \#2. Protocol \#1 is the mean per-joint position error (MPJPE) in millimeters between the predicted joint positions and the ground-truth joint positions. Protocol \#2 computes the same error, but after applying a subsequent rigid transformation that further aligns the predictions with the ground truth, and is alternately referred to as P-MPJPE in the literature. 

\subsection{Hypothesis selection}
The MDN outputs are distributions as opposed to most approaches on Human3.6M, which only predict a single scalar. To compare with previous approaches we need to reduce the distributions to a single prediction. In this paper we use the following methods for creating a prediction:

\textbf{Highest}: Simply choose the mean of the kernel $n$ with the highest mixture coefficient. $\hat{\Vec{y}} = \mu_n \mid n = \text{argmax}_j\pi_j$. This essentially represents the guess the network thinks is the most likely to be correct.

\textbf{Mean}: Predict the weighted average of the distribution calculated as 
$\hat{\Vec{y}} = \sum_{j=1}^K \pi_j \mu_j$.
This is useful because of the way error is evaluated. If we know the true distribution of target variables, in order to minimize average MPJPE-error it is beneficial to predict the mean of the distribution, even if the mean itself is not a possible pose. 

\textbf{Oracle}: Since each kernel represents a distinct possible pose, having an \textit{Oracle} to predict which one of these possible poses to choose can improve our results significantly. The \textit{Oracle} selects out of the K kernel means the 3D pose that is closest to the target 3D pose. Previous works treating pose estimation as an inverse problem take this approach \cite{Li_2019_CVPR,jahangiri2017generating, Sharma_2019_ICCV}, and it can be considered as an upper-bound estimate for how well the GraphMDN is performing on this task. While this is not a realistic method for real-world applications (we would never need a prediction mechanism in the first place if we knew the ground truth target), we nevertheless make use of it in order to compare against alternative multi-modal methods.

\subsection{Training details}

In this work we aimed to keep our training hyperparameters and setup as close to SemGCN as possible. However, in order to speed up the training process itself, we increased batch-size from 64 to 256 and replaced the decaying learning rate with the \textit{Super-Convergence} LR schedule described in \cite{smith2017superconvergence} with a peak learning rate of $6 \times 10^{-3}$. This allowed us to train all networks in just 2 epochs instead of 30+ used by \cite{Zhao_2019_CVPR}. Together these changes allow us to reduce training times from 10 hours to less than 15 minutes using the same hardware. We also reproduce the SemGCN results using Super-Convergence and our setup to isolate the effects of the new training setup from the effects of the MDN output. All our models used 5 MDN kernels unless otherwise stated. We train a single model for all actions. Our model is implemented in PyTorch, and we employ ADAM \cite{kingma2014adam} for optimization and initialize our network weights using Xavier initialization \cite{glorot2010understanding}. All models used dropout with a rate of 0.1 unless otherwise mentioned.

\section{Results}
\label{results}

\begin{table}
\scriptsize
\setlength{\tabcolsep}{1pt}
\begin{tabular}{lllllllllllllllll}
\noalign{\smallskip}
\textbf{Protocol   \#1} & Direct. & Discuss & Eating & Greet & Phone & Photo & Posing & Purch. & Sitting & SittingD. & Smoke & Wait & WalkD & Walk & WalkT. & Avg. \\ \hline
SemGCN (reported)\cite{Zhao_2019_CVPR} & 37.8 & 49.4 & 37.6 & 40.9 & 45.1 & 41.4 & 40.1 & 48.3 & 50.1 & 42.2 & 53.5 & 44.3 & 40.5 & 47.3 & 39.0 & 43.8 \\
SemGCN & 34.7 & 41.9 & 35.3 & 38.7 & 41.0 & 53.5 & 41.1 & 37.3 & 44.6 & 53.9 & 40.3 & 41.9 & 41.5 & 33.3 & 36.0 & 41.0 \\
SemGCN (Wide) & 34.4 & 41.6 & 32.9 & 37.5 & 39.2 & 47.5 & 41.0 & 34.1 & 43.6 & 52.3 & 37.6 & 40.7 & 38.6 & 30.5 & 32.9 & 39.0 \\
Ours (Mean) & \textbf{32.2} & \textbf{39.5} & 33.5 & 36.7 & 38.2 & 48.5 & 39.0 & 36.7 & 44.0 & 52.9 & 37.2 & 40.3 & 39.4 & 30.0 & 32.1 & 38.7 \\
Ours (Highest, Wide) & 34.5 & 40.5 & \textbf{33.0} & 35.7 & 37.0 & 44.8 & 39.1 & \textbf{33.0} & 41.2 & 50.2 & 36.6 & 38.6 & 38.2 & 28.4 & 31.8 & 37.5 \\
Ours (Mean, Wide) & 33.9 & 39.9 & \textbf{33.0} & \textbf{35.4} & \textbf{36.8} & \textbf{44.4} & \textbf{38.9} & \textbf{33.0} & \textbf{41.0} & \textbf{50.0} & \textbf{36.4} & \textbf{38.3} & \textbf{37.8} & \textbf{28.2} & \textbf{31.5} & \textbf{37.2} \\ \hline
Ours (Oracle, Wide) & 28.9 & 34.5 & 28.2 & 30.2 & 31.5 & 38.5 & 32.3 & 28.6 & 35.7 & 43.3 & 31.9 & 32.1 & 33.3 & 25.2 & 27.8 & 31.8 \\
 &  &  &  &  &  &  &  &  &  &  &  &  &  &  &  &  \\
\textbf{Protocol \#2} & Direct. & Discuss & Eating & Greet & Phone & Photo & Posing & Purch. & Sitting & SittingD. & Smoke & Wait & WalkD & Walk & WalkT. & Avg. \\ \hline
SemGCN & 26.2 & 32.7 & 28.4 & 31.4 & 31.3 & 40.8 & 31.0 & 28.6 & 35.9 & 43.5 & 32.0 & 32.2 & 33.2 & 26.8 & 29.2 & 32.2 \\
SemGCN (Wide) & 26.4 & 32.4 & 28.0 & 30.2 & 30.9 & 36.9 & 31.5 & 26.9 & 36.1 & 41.6 & 30.8 & 31.6 & 31.3 & 24.3 & 27.1 & 31.1 \\
Ours (Mean) & \textbf{24.6} & \textbf{30.5} & \textbf{26.3} & 29.7 & \textbf{28.8} & 37.0 & 30.1 & 26.8 & \textbf{34.0} & 40.8 & \textbf{29.5} & 31.5 & 31.0 & 24.2 & \textbf{25.9} & 30.0 \\
Ours (Highest, Wide) & 25.8 & 31,4 & 27.5 & 28.7 & 29.1 & 35.2 & 29.8 & 26.0 & 34.2 & \textbf{40.5} & 29.8 & 30.1 & 30.4 & \textbf{22.3} & 26.0 & 29.8 \\
Ours (Mean, Wide) & 25.4 & 30.9 & 27.5 & \textbf{28.6} & 29.1 & \textbf{35.0} & \textbf{29.7} & \textbf{25.9} & 34.1 & \textbf{40.5} & 29.8 & \textbf{29.9} & \textbf{30.2} & 22.5 & \textbf{25.9} & \textbf{29.7} \\ \hline
Ours (Oracle, Wide) & 22.6 & 27.8 & 23.9 & 24.9 & 26.0 & 31.4 & 25.6 & 22.8 & 30.3 & 35.9 & 26.8 & 26.3 & 26.9 & 20.2 & 22.2 & 26.3 \\
 &  &  &  &  &  &  &  &  &  &  &  &  &  &  &  & 
\end{tabular}
\caption{(P) MPJPE in millimeter on Human3.6M under protocol \#1 and \#2 using the ground-truth 2D joint positions as inputs. We can see our GCMDN consistently outperforms SemGCN using comparable hypothesis selection mechanisms (Mean, Highest) and clearly beats it using Oracle which highlights the usefulness of our multimodal approach.}
\label{tab:GT_ours_semgcn}

\end{table}

\begin{table}
\scriptsize
\setlength{\tabcolsep}{1pt}
\begin{tabular}{lcccccccccccccccc}
\textbf{Protocol   \#1} & Direct. & Discuss & Eating & Greet & Phone & Photo & Posing & Purch. & Sitting & SittingD. & Smoke & Wait & WalkD & Walk & WalkT. & Avg. \\ \hline
SemGCN (reported)\cite{Zhao_2019_CVPR} & \textbf{48.2} & 60.8 & \textbf{51.8} & 64.0 & 64.6 & 53.6 & \textbf{51.1} & 67.4 & 88.7 & \textbf{57.7} & 73.2 & 65.6 & \textbf{48.9} & 64.8 & 51.9 & 60.8 \\
SemGCN & 53.7 & 57.9 & 58.0 & 59.1 & 66.0 & 78.1 & 55.7 & 58.1 & 71.7 & 96.1 & 62.8 & 60.6 & 65.4 & 53.3 & 57.3 & 63.6 \\
SemGCN (Wide) & 49.6 & 55.1 & 56.5 & \textbf{55.9} & 62.4 & 74.3 & 51.4 & 54.6 & 69.8 & 92.6 & 59.6 & 56.9 & 60.5 & \textbf{48.3} & 52.1 & 60.0 \\
Ours (Mean) & 51.9 & 56.1 & 55.3 & 58.0 & 63.5 & 75.1 & 53.3 & 56.5 & 69.4 & 92.7 & 60.1 & 58.0 & 65.5 & 49.8 & 53.6 & 61.3 \\
Ours (Mean, Wide) & 49.9 & \textbf{54.9} & 55.2 & 56.0 & \textbf{62.1} & \textbf{73.2} & 51.6 & \textbf{53.2} & \textbf{69.0} & 88.2 & \textbf{58.9} & \textbf{55.8} & 61.0 & 48.6 & \textbf{50.1} & \textbf{59.2} \\
 &  &  &  &  &  &  &  &  &  &  &  &  &  &  &  &  \\
\textbf{Protocol \#2} & Direct. & Discuss & Eating & Greet & Phone & Photo & Posing & Purch. & Sitting & SittingD. & Smoke & Wait & WalkD & Walk & WalkT. & Avg. \\ \hline
SemGCN & 40.7 & 44.7 & 45.6 & 46.5 & 50.6 & 56.4 & 40.4 & 41.8 & 57.3 & 71.1 & 49.8 & 45.6 & 50.1 & 40.8 & 44.4 & 48.4 \\
SemGCN (Wide) & 38.9 & 42.9 & 44.5 & \textbf{44.8} & 48.8 & 54.5 & \textbf{39.0} & 39.9 & 56.0 & 68.4 & 47.9 & 43.1 & 47.0 & \textbf{36.8} & 41.8 & 46.3 \\
Ours (Mean) & 39.7 & 43.4 & \textbf{44.0} & 46.2 & 48.8 & 54.5 & 39.4 & 41.1 & 55.0 & 69.0 & 48.0 & 43.7 & 49.6 & 38.4 & 42.4 & 46.9 \\
Ours (Mean, Wide) & \textbf{38.5} & \textbf{42.6} & 44.1 & 44.9 & \textbf{48.1} & \textbf{53.3} & \textbf{39.0} & \textbf{39.5} & \textbf{54.9} & \textbf{66.2} & \textbf{47.0} & \textbf{42.2} & \textbf{46.8} & \textbf{36.8} & \textbf{39.8} & \textbf{45.6} \\
 &  &  &  &  &  &  &  &  &  &  &  &  &  &  &  & 
\end{tabular}
\label{tab:SH_ours_semgcn}
\caption{(P) MPJPE in millimeter on Human3.6M under protocol \#1 and \#2 using the fine-tuned stacked hourglass 2D inputs. Our method outperforms SemGCN using SH inputs.}
\end{table}

\subsection{Comparison to SemGCN}

In Table 1 we compare our method against SemGCN \cite{Zhao_2019_CVPR} on ground truth (GT) 2D inputs, and in Table 2 we do the same comparisons using Stacked Hourglass (SH) inputs. All methods were trained in 2 epochs using Super-Convergence. Our standard models used a set of hyperparameters matching the SemGCN architecture, with 4 block layers having 128-dimensional hidden layers. Additionally we experimented with a wider network using 3 blocks with a hidden dimension of 512; these results are reported are indicated as (Wide). Regardless of the input type (SH or GT), we can see that our GraphMDN noticeably improves over SemGCN, such that simply replacing SemGCN's output layer with a GraphMDN output reduces average P1-error by up to 5\%. Furthermore, GraphMDN learns a richer representation of the data capable of furnishing predictions about the \textit{set} of poses that reduce to a particular 2D input. It worth noting that our \textit{Oracle} results are substantially better than other published methods using single frame GT inputs, indicating that refinement of hypothesis-selection procedures could yield significant improvement in real-world applications and represents a promising direction for future research.

\subsection{Comparison against multimodal methods}

To evaluate the quality of our learned distributions, we compare our models against previous multimodal approaches of human pose estimation \cite{jahangiri2017generating, Li_2019_CVPR,Sharma_2019_ICCV} using an \textit{Oracle} to choose the best prediction. The results are shown in Table 3. We can see our method produces state of the art results with 5 guesses after only 2 epochs of training. Additionally our MDN scales much better to larger numbers of kernels than \cite{Li_2019_CVPR}, with a significant performance gain when going from 5 to 8 kernels, while \cite{Li_2019_CVPR} does not gain performance beyond 5 kernels. In addition, our method scales well even up to 200 kernels, and is able to outperform \cite{Sharma_2019_ICCV} on a best of 200 predictions under Protocol 1, although the method in \cite{Sharma_2019_ICCV}
still outperforms GraphMDN under Protocol 2. When GraphMDN is trained with 5 kernels, it effectively generates 5 poses as predictions of the underlying 3D pose. Compared to the accuracy of 5 samples (poses) from the latent space generated in \cite{Sharma_2019_ICCV}, GraphMDN clearly outperforms the variational autoencoder approach in \cite{Sharma_2019_ICCV}. When training the model with 200 kernels, we found Super-Convergence to reduce \textit{Oracle} performance; we hypothesize that reduced performance is due to the rapid training procedure reducing kernel diversity. Because of this, our 200 kernel model was trained with 30 epochs using an exponentially decaying learning rate starting at $10^{-3}$ and a dropout rate of 0.5. While some unimodal approaches have attained better results on this task \cite{Iskakov_triangulation, pavllo20193d, Qiu_2019_ICCV}, none of the outperforming methods are comparable as they often utilize either multi-view or time-series data, or 2D joint detectors better than Stacked Hourglass. An obvious source of improvement would be to utilize a more modern 2D joint detector such as CPN\cite{pavllo20193d}, which would clearly improve our results, but this would also improve the results of other multimodal methods and as such is not meaningful when comparing against them.

\begin{table}
\scriptsize
\setlength{\tabcolsep}{1pt}
\begin{tabular}{lllllllllllllllll}
\textbf{Protocol   \#1} & Direct. & Discuss & Eating & Greet & Phone & Photo & Posing & Purch. & Sitting & SittingD. & Smoke & Wait & WalkD & Walk & WalkT. & Avg. \\ \hline
Li et al.\cite{Li_2019_CVPR}(5) & \textbf{43.8} & \textbf{48.6} & 49.1 & 49.8 & 57.6 & \textbf{61.5} & 45.9 & 48.3 & 62 & \textbf{73.4} & 54.8 & 50.6 & 56 & \textbf{43.4} & 45.5 & 52.7 \\
Sharma et al.\cite{Sharma_2019_ICCV}(5)* & - & - & - & - & - & - & - & - & - & - & - & - & - & - & - & 55.4 \\
Ours (Wide, 5) & 44.2 & 48.7 & \textbf{47.2} & \textbf{49.4} & \textbf{55.1} & 62.0 & \textbf{44.8} & \textbf{46.9} & \textbf{59.7} & 76.5 & \textbf{53.2} & \textbf{48.9} & \textbf{53.6} & 44.5 & \textbf{44.4} & \textbf{51.9} \\ \hline
Li et al.\cite{Li_2019_CVPR}(8) & - & - & - & - & - & - & - & - & - & - & - & - & - & - & - & 52.6 \\
Ours (8) & 43.4 & 47.4 & 46.3 & 48.1 & 55.2 & 59.3 & 43.9 & 45.8 & 58.6 & 75.2 & 52.5 & 48.2 & 53.2 & 42.2 & 43.6 & \textbf{50.9} \\ \hline
Sharma et al.\cite{Sharma_2019_ICCV}(200) & \textbf{37.8} & \textbf{43.2} & 43.0 & 44.3 & 51.1 & 57.1 & \textbf{39.7} & 43.0 & 56.3 & \textbf{64.0} & \textbf{48.1} & 45.4 & 50.4 & \textbf{37.9} & \textbf{39.9} & 46.8 \\
Ours (Wide, 200) & 40.0 & \textbf{43.2} & \textbf{41.0} & \textbf{43.4} & \textbf{50.0} & \textbf{53.6} & 40.1 & \textbf{41.4} & \textbf{52.6} & 67.3 & \textbf{48.1} & \textbf{44.2} & \textbf{49.0} & 39.5 & 40.2 & \textbf{46.2} \\
 &  &  &  &  &  &  &  &  &  &  &  &  &  &  &  &  \\
\textbf{Protocol \#2} & Direct. & Discuss & Eating & Greet & Phone & Photo & Posing & Purch. & Sitting & SittingD. & Smoke & Wait & WalkD & Walk & WalkT. & Avg. \\ \hline
Li et al.\cite{Li_2019_CVPR}(5) & 35.5 & 39.8 & 41.3 & 42.3 & 46.0 & 48.9 & 36.9 & 37.3 & 51.0 & 60.6 & 44.9 & 40.2 & 44.1 & \textbf{33.1} & 36.9 & 42.6 \\
Ours (Wide, 5) & \textbf{33.8} & \textbf{38.7} & \textbf{38.3} & \textbf{39.4} & \textbf{43.6} & \textbf{47.8} & \textbf{34.2} & \textbf{35.2} & \textbf{48.5} & \textbf{58.6} & \textbf{42.7} & \textbf{37.8} & \textbf{41.4} & 33.5 & \textbf{34.4} & \textbf{40.5} \\ \hline
Ours (8) & 33.8 & 38.3 & 37.8 & 38.8 & 43.8 & 46.3 & 33.9 & 34.6 & 47.9 & 58.6 & 42.4 & 37.6 & 41.1 & 32.0 & 34.2 & 40.1 \\ \hline
Sharma et al.\cite{Sharma_2019_ICCV}(200) & \textbf{27.6} & \textbf{27.5} & 34.9 & \textbf{32.3} & \textbf{33.3} & 42.7 & \textbf{28.7} & \textbf{28.0} & \textbf{36.1} & \textbf{42.7} & \textbf{36.0} & \textbf{30.7} & \textbf{37.6} & \textbf{24.3} & \textbf{27.1} & \textbf{32.7} \\
Ours (Wide, 200) & 30.8 & 34.7 & \textbf{33.6} & 34.2 & 39.6 & \textbf{42.2} & 31.0 & 31.9 & 42.9 & 53.5 & 38.1 & 34.1 & 38.0 & 29.6 & 31.1 & 36.3 \\
 &  &  &  &  &  &  &  &  &  &  &  &  &  &  &  & 
\end{tabular}
\label{tab:oracle}
\caption{Comparison of state of the art multimodal methods evaluated under the Oracle protocol. All models use the stacked hourglass 2D inputs. * was estimated from figure. The number in brackets is the number of pose predictions considered; i.e. the number of Gaussian kernels (our paper and \cite{Li_2019_CVPR}) or samples generated \cite{Sharma_2019_ICCV}.}
\end{table}

\subsection{Qualitative Analysis of GraphMDNs}

\begin{figure*}[h]
  \begin{center}
  %\framebox[4.0in]{$\;$}
  \includegraphics[width=1.0\linewidth]{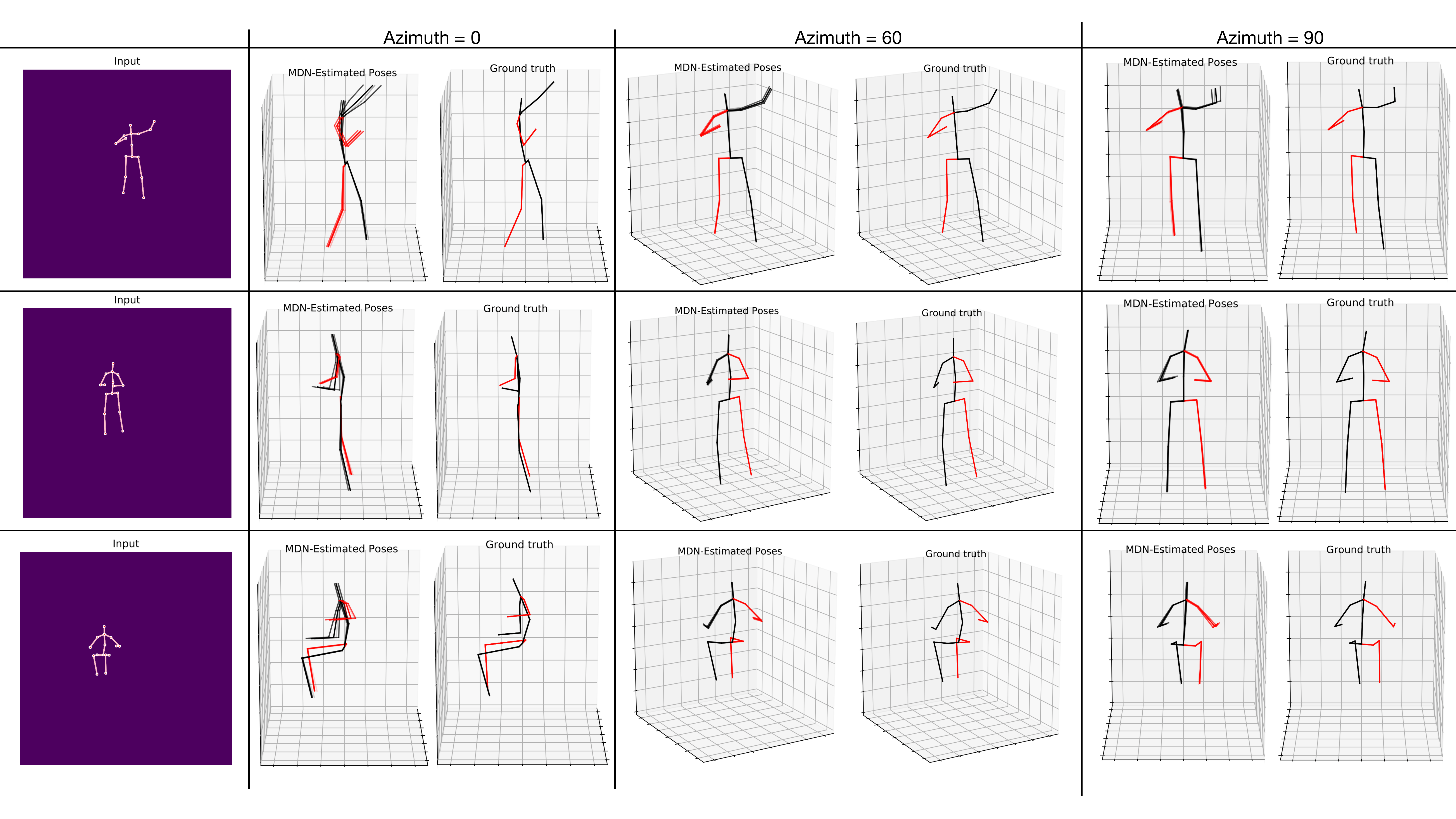}
  \end{center}
  \caption{Qualitative results on the Human36M data set using ground-truth 2D inputs. The left column, labeled ``Input'', displays the input (2D) skeleton. The next three columns compare the output of our GraphMDN (labeled ``MDN-Estimated Poses'') with the ground truth target value (labeled "Ground truth") at three different camera azimuths (labeled on figure). In each plot of our MDN-Estimated Poses, each skeleton is shaded according to the mixing coefficient of the kernel that generated it -- kernels with low mixing coefficients will be nearly transparent, while kernels with high mixing coefficients will be opaque. From top row to bottom row, the ``Greeting'', ``Purchasing'', and ``Sitting'' actions are displayed.}
  \label{fig:occlusion_figure}
\end{figure*}

Although the \textit{Oracle} predictions are, on their own, not a plausible mechanism for making predictions in the real-world, they nevertheless indicate that representing multiple potential alternatives can \textit{in principle} result in great improvements in performance. Indeed, the \textit{Oracle} result from the GraphMDN model far exceeds the performance of the SemGCN model on its own, and outperforms state-of-the-art \textit{Oracle} predictions from alternative multi-modal modelling techniques. An obvious question, then, is what exactly have the various kernels in our predictions learned to represent, and subsequently, is there a ``fair''  mechanism that would allow selection between the various hypotheses (without requiring foreknowledge of the answer)?

In order to understand how the mixture model functions in our predictions, Figure 2 shows visualizations of the predicted kernels (super-imposed on one another, with shading proportional to their weighting coefficients) selected from frames in which the best performing kernel performs significantly better than the weighted mean of the predicted kernels. When we visualize the 2D inputs used to make the 3D joint predictions, it is immediately clear that there is ambiguity in the depth axis of the camera, as to where exactly some of the joints might be located. In fact, when viewing the 3D predictions from the camera angle used to generate the 2D inputs (azimuth=60; the middle column for each of the rows of Figure 2 the multiple hypotheses are sufficiently indistinguishable that they often simply appear to be a single prediction. Nevertheless, when rotating that view (the columns corresponding to Azimuth values of 0 and 90), it becomes clear that the the multiple-hypotheses all correspond to distinct joint positions along the depth axis of the camera.

As demonstrated by the feasibility of all hypotheses output by the GraphMDN (given the 2D inputs), the mixture model has learned to meaningfully represent the statistics of the output distribution. Moreover, the fact that the model is capable of providing multiple hypotheses allows for additional information to be applied when deciding which of the possible kernels represents the best predicted 3D joint locations. In this case, additional information can include anything from human-kinematic structural constraints to multi-view information as well as temporal information (for videos). Although unconstrained optimization over this information can be a potentially costly process, selecting the best of N hypotheses can often greatly simplify the solution search space. While the present GraphMDN work does not make use of these constraints in selecting between alternative hypotheses, the visual confirmation that the distribution of 3D-joint hypotheses represent alternatives that cannot be distinguished by a human observer from the 2D image alone, provides further evidence that the combined GNN + MDN approach results in useful representations that would be difficult to achieve with only one of the other techniques.